\documentclass{article}
\usepackage{spconf,amsmath,amssymb,graphicx}
\usepackage{booktabs}
\usepackage{xurl}
\usepackage{multirow}
\usepackage{array}
\usepackage[hidelinks]{hyperref}
\title{Zero-Shot Cue-Grounded Topic Segmentation of Spoken Documents}

\name{Suhwan Choi$^{1}$, Myeongho Jeon$^{2}$, Myungjoo Kang$^{1}$}
\address{$^{1}$Seoul National University, $^{2}$KAIST}
\hypersetup{pdfauthor={Suhwan Choi, Myeongho Jeon, Myungjoo Kang}}

\begin{document}
\ninept
\maketitle

\begin{abstract}
Topic segmentation structures spoken documents into coherent sections, facilitating navigation and downstream understanding. The appropriate granularity can vary substantially, ranging from broad thematic shifts to fine-grained subtopics. Existing LLM-based segmenters, however, often struggle to adapt to this variation, causing them to either merge distinct subtopics or over-segment coherent themes. To address this, we introduce \textbf{C}ue-\textbf{G}rounded \textbf{S}egmentation (CGS), a training-free framework that operates without any task-specific supervision. CGS first identifies phrases that explicitly signal the start of a new topic and uses their sentence positions as segment boundaries. When such cues are insufficient, it falls back to semantic segmentation, guided by the document structure inferred during cue extraction. Across six benchmarks and six LLM backbones, CGS consistently outperforms existing baselines, remains robust to noisy ASR transcripts, and achieves these gains with low API cost on proprietary models.

\end{abstract}


\section{Introduction}
\label{sec:intro}

Topic segmentation organizes spoken documents into coherent sections by identifying where new topics begin. These topic-level units are essential for efficiently navigating, retrieving, and summarizing long-form recordings. Unlike written documents, however, spoken content rarely provides explicit structural cues such as headings or section titles, requiring topic boundaries to be inferred from the discourse itself, including transition phrases and shifts in subject matter.

Classical unsupervised methods place boundaries where lexical cohesion drops \cite{hearst1997texttiling,eisenstein2008bayesian}, and speech segmentation methods use prosody and discourse markers \cite{shriberg2000prosody,galley2003lcseg}. Representation-based methods learn boundary decisions from labeled corpora \cite{lukasik2020text,lo2021transformer}, train per-document segment representations without labels \cite{wang2023m3seg}, or hierarchically cluster transcript embeddings \cite{augmend2024treeseg}. However, these methods determine section granularity through thresholds, supplied segment counts, or learned boundary rules, which can limit generalization across diverse document structures.
Recent LLM systems predict topic boundaries through boundary lists, sentence-level topic labels, recursive splitting, table-of-contents planning, topic-shift predictions based on utterance intent, or iterative chunking \cite{segmentllm2024,mackenzie2025llmsegmentation,freisinger2025toclm,duarte2024lumberchunker,defdts2025}.
These methods directly predict boundaries or segment structures \textit{without explicitly grounding them} in topic-opening cues. To make these predictions, the model must decide whether each change in the discussion marks a new topic or is only a minor shift within the current topic. Treating minor shifts as new topics leads to over-segmentation, while overlooking meaningful topic changes leads to under-segmentation.

\begin{figure}[!t]
\centering
\includegraphics[width=\columnwidth]{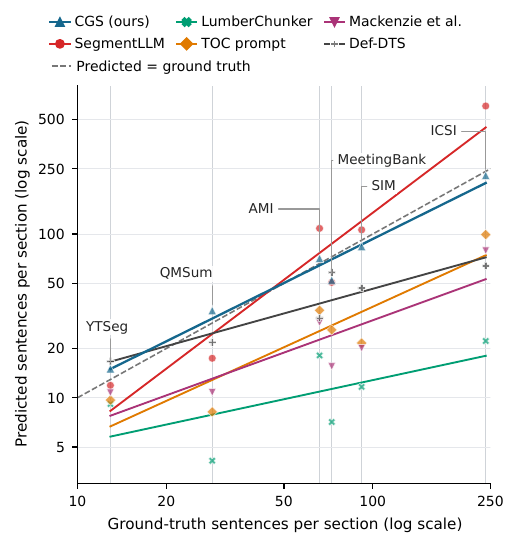}
\caption{Predicted versus ground-truth section length. CGS more closely tracks ground-truth lengths across corpora, while baselines often produce sections that are too short or too long. Lengths are defined in Table~\ref{tab:corpora}; predictions are averaged across six backbones. Vertical lines mark corpora; colored lines show log--log fits.}
\label{fig:motivation}
\end{figure}
\begin{table}[!t]
\centering
\caption{Evaluated corpus subsets. $n$ counts documents. For each document, we calculate the average ground-truth segment length in sentences. Length reports the median across documents.}
\label{tab:corpora}
\vspace{3pt}
\begin{tabular}{@{}l@{\hspace{5pt}}r@{\hspace{5pt}}r@{\hspace{7pt}}l@{}}
\toprule
Dataset & $n$ & Length & Description \\
\midrule
YTSeg~\cite{retkowski2024ytseg} & 1,448 & 12.9 & Creator-authored chapters \\
\addlinespace[3pt]
ICSI~\cite{janin2003icsi,gu2024tcr} & 75 & 241.0 & Research meetings \\
\addlinespace[3pt]
AMI~\cite{carletta2007ami} & 129 & 66.0 & Multi-party meetings \\
\addlinespace[3pt]
MeetingBank~\cite{hu2023meetingbank,gu2024tcr} & 28 & 72.6 & City-council meetings \\
\addlinespace[3pt]
QMSum~\cite{zhong2021qmsum,gu2024tcr} & 20 & 28.6 & Parliamentary committees \\
\addlinespace[3pt]
SIM~\cite{gu2024tcr} & 100 & 91.6 & Spliced meeting excerpts \\
\bottomrule
\end{tabular}
\end{table}

\begin{figure*}[t]
\centering
\includegraphics[width=\textwidth]{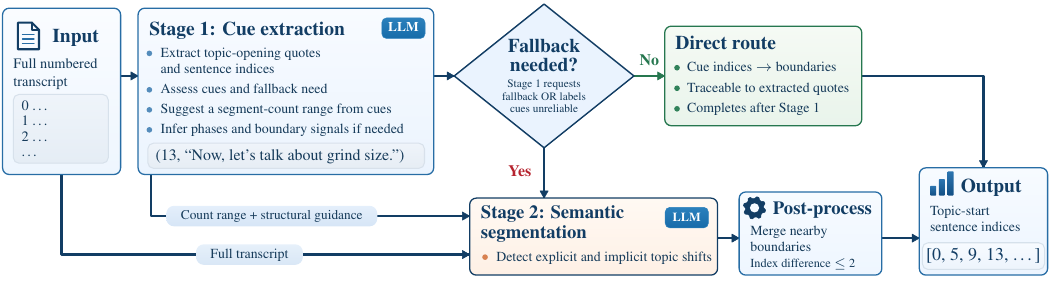}
\caption{CGS overview. Each cue pairs a topic-opening
quote with its sentence index. Both stages use the same LLM with task-specific
prompts. Structural guidance covers phases, boundary signals, and expected
segment durations.}
\label{fig:pipeline}
\end{figure*}

Spoken documents vary substantially in their topical structure. Some consist of short, clearly separated topics, whereas others sustain extended discussions around a single subject. An effective topic segmenter must therefore adapt to a wide range of segment granularities. This variation is evident across six benchmarks. The median ground-truth segment length ranges from 12.9 sentences in chaptered YouTube videos to 241 sentences in research-group meetings, an 18.6-fold difference. Despite this substantial variation, existing LLM-based segmenters often produce segments that are systematically too short or too long for the target corpus (Fig.~\ref{fig:motivation}).

\begingroup\clubpenalty=10000
In this regard, we propose \underline{\textbf{C}}ue-\underline{\textbf{G}}rounded \underline{\textbf{S}}egmentation (CGS)\footnote{Full prompts, output schemas, and code are available at\\{\urlstyle{same}\url{https://github.com/schoi828/CGS}}.}, a zero-shot framework that grounds boundary prediction in topic-opening cues. Such cues can indicate which parts of a discussion are presented as distinct topics. When these cues are sufficiently informative, CGS uses their sentence positions as boundaries, allowing section lengths to reflect the spacing between topic openings. When cue evidence is insufficient, CGS falls back to semantic segmentation guided by the document structure inferred during cue extraction. This design enables CGS to better adapt to the wide variation in ground-truth segment lengths across the six benchmarks (Fig.~\ref{fig:motivation}).\par
\endgroup

More specifically, CGS is implemented in two stages (Fig.~\ref{fig:pipeline}). Stage~1 reads the full numbered
transcript, quotes topic-opening phrases, and reports their sentence
indices. On the direct route, those indices define the boundaries, so
each cue-derived boundary is paired with the transcript phrase that
motivated it. When Stage~1 requests fallback or judges the cues unreliable,
Stage~2 re-reads the full transcript, guided by the expected topic count,
structural phases, and boundary signals inferred in Stage~1. It identifies topic changes from the transcript's semantic content, without explicit transition phrases.

Our main contributions and findings are:
\begin{list}{\labelitemi}{%
  \setlength{\topsep}{3pt}%
  \setlength{\itemsep}{1pt}%
  \setlength{\parsep}{0pt}%
  \setlength{\partopsep}{0pt}%
  \def\makelabel#1{\hss\llap{#1}}}
\item We introduce CGS, a training-free, zero-shot method that uses quoted topic openings to locate boundaries and invokes structure-guided semantic segmentation when those cues are insufficient.
\item Across six benchmarks and six LLM backbones, CGS achieves the best average performance on all metrics, with low API cost on proprietary models.
\item Ablations show the benefit of combining cue extraction with semantic fallback and supplying segment-count guidance. Automatic speech recognition (ASR) experiments show that CGS retains its performance advantage under speech recognition noise.
\end{list}

\section{Methodology}
\label{sec:method}

CGS consists of two stages. First, it identifies explicit transition cues in the transcript and determines whether they provide sufficient evidence for topic boundaries. When such cues are insufficient, CGS falls back to semantic segmentation guided by the document structure inferred during cue extraction (Fig.~\ref{fig:pipeline}). The two stages are implemented as separate calls to the same LLM. We use a single shared set of prompts across all datasets, without providing dataset names, task-specific tuning, or labeled examples.

\subsection{Stage 1: Indexed Transition Evidence}
\label{sec:stage1}

Stage~1 processes the full transcript and returns a structured JSON output. The prompt asks the model to identify and quote each topic-opening phrase together with its sentence index, forming the cue list $C$. In the same response, the model assigns an overall cue-reliability label---reliable, uncertain, or unreliable---and indicates whether semantic fallback is required. Stage~2 is invoked when fallback is explicitly requested or when the cue list is labeled unreliable; an uncertain label alone does not trigger fallback.

\begin{table*}[!t]
  \centering
  \scriptsize
  \setlength{\tabcolsep}{2.4pt}
  \caption{Performance on the six benchmark corpora. \textbf{Bold} and \underline{underline} mark the best and second-best values per column. $^{\dagger}$Training-free, but uses self-supervised BERT pretraining. $^{\ddagger}$Def-DTS uses only valid outputs completed within the output-token limit of backbones.}
  \label{tab:main}
\vspace{3pt}
  \begin{tabular*}{\textwidth}{@{\extracolsep{\fill}}ll ccccccc ccccccc ccccccc@{}}
    \toprule
    & & \multicolumn{7}{c}{$P_k$ ($\downarrow$)}
    & \multicolumn{7}{c}{WindowDiff ($\downarrow$)}
    & \multicolumn{7}{c}{Boundary $F_1$ ($\uparrow$)} \\
    \cmidrule(lr){3-9} \cmidrule(lr){10-16} \cmidrule(l){17-23}
    & Method & YT & IC & MB & SIM & AMI & QM & Avg.
           & YT & IC & MB & SIM & AMI & QM & Avg.
           & YT & IC & MB & SIM & AMI & QM & Avg. \\
    \midrule
    \multirow{2}{*}{Classical} &
    \textsc{TextTiling}~\cite{hearst1997texttiling} &
      .513 & .716 & .648 & .604 & .611 & .630 & .620 &
      .653 & .996 & .994 & 1.00 & .816 & .999 & .910 &
      .433 & .036 & .086 & .057 & .121 & .129 & .144 \\
    & \textsc{BERT-TT}~\cite{solbiati2021unsupervised}$^{\dagger}$ &
      .402 & .473 & .420 & .378 & .420 & \underline{.407} & .417 &
      .409 & .514 & .464 & .434 & .442 & \underline{.432} & .449 &
      .189 & .055 & .133 & .197 & .102 & .188 & .144 \\
    \midrule
    \multirow{5}{*}{\shortstack[l]{LLM-based}} &
    \textsc{LumberChunker}~\cite{duarte2024lumberchunker} &
      .378 & .714 & .641 & .604 & .540 & .623 & .583 &
      .473 & .983 & .952 & .994 & .732 & .972 & .851 &
      .539 & .121 & .182 & .119 & .291 & .233 & .247 \\
    & Mackenzie et al.~\cite{mackenzie2025llmsegmentation} &
      .349 & .533 & .460 & .540 & .459 & .511 & .476 &
      .421 & .722 & .692 & .844 & .599 & .712 & .665 &
      .519 & .182 & .293 & .141 & .302 & .226 & .277 \\
    & \textsc{TOC prompt}~\cite{freisinger2025toclm} &
      .414 & .516 & .472 & .574 & .511 & .495 & .497 &
      .483 & .632 & .648 & .840 & .606 & .693 & .650 &
      .453 & .098 & .181 & .090 & .158 & .322 & .217 \\
    & \textsc{Def-DTS}$^{\ddagger}$~\cite{defdts2025} &
      .338 & .400 & .287 & .317 & .304 & .406 & .340 & .368 & .490 & .402 & .430 & .387 & .536 & .433 & .325 & .170 & .272 & .279 & .358 & .251 & .289 \\
    & \textsc{SegmentLLM}~\cite{segmentllm2024} &
      \underline{.271} & \underline{.333} & \underline{.206} & \underline{.306} & \underline{.309} & .416 & \underline{.307} &
      \underline{.324} & \underline{.402} & \underline{.275} & \underline{.370} & \underline{.367} & .535 & \underline{.379} &
      \underline{.590} & \underline{.200} & \underline{.618} & \underline{.350} & \underline{.330} & \underline{.382} & \underline{.412} \\
    \midrule
    Ours & \textbf{CGS} & \textbf{.245} & \textbf{.212} & \textbf{.069} & \textbf{.206} & \textbf{.216} & \textbf{.301} & \textbf{.208} & \textbf{.274} & \textbf{.272} & \textbf{.155} & \textbf{.241} & \textbf{.269} & \textbf{.333} & \textbf{.257} & \textbf{.604} & \textbf{.395} & \textbf{.751} & \textbf{.471} & \textbf{.438} & \textbf{.391} & \textbf{.508} \\
    \bottomrule
  \end{tabular*}
\end{table*}

\begingroup\widowpenalty=10000
Stage~1 also predicts a segment-count range from the extracted cues. Given $n$ cues, the lower and upper bounds are set to $\max(1,n-1)$ and $n+1$, respectively. When fallback is required, the model additionally produces structural guidance for Stage~2: phases summarizing the document's major parts and their order, boundary signals describing likely indicators of topic transitions, and a duration range estimating plausible segment lengths in seconds from the transcript. When fallback is not required, these fields are omitted. Table~\ref{tab:guidance_example} illustrates the Stage~1 output using an example from an AMI meeting.\par
\endgroup

\begingroup\interlinepenalty=10000
The direct route completes segmentation after Stage~1.
CGS discards out-of-range cue indices and returns
$B_{\mathrm{cue}}=\{0\}\cup\{i:(i,q)\in C\}$ over the remaining pairs.\par
\endgroup

\subsection{Stage 2: Conditioned Semantic Fallback}
\label{sec:routing}

Stage~2 reads the full numbered transcript together with the segment-count range and structural guidance from Stage~1. The LLM uses this guidance to identify coherent sections and places boundaries where the discussion moves to a new topic. It can identify these transitions without an explicit phrase such as ``Next, let's discuss staffing.'' Stage~2 predicts a new set of topic-start sentence indices from the transcript; it does not receive the Stage-1 cue list. The suggested segment-count range guides this prediction rather than imposing a strict constraint. To reduce the number of very short segments, CGS merges neighboring
Stage~2 boundary predictions whose sentence indices differ by at most two.

\begin{table}[!ht]
\centering
\fontsize{9}{10.8}\selectfont
\setlength{\tabcolsep}{3pt}
\caption{Stage-1 guidance supplied to Stage~2: an example from an AMI meeting. Phase and signal entries are excerpts; both ranges are shown in full.}
\label{tab:guidance_example}
\vspace{3pt}
\begin{tabular}{@{}>{\raggedright\arraybackslash}p{.30\columnwidth}>{\raggedright\arraybackslash}p{.65\columnwidth}@{}}
\toprule
Guidance & Saved example \\
\midrule
Segment count & 3--5 segments \\
\addlinespace[2pt]
Structural phases & \emph{Prototype Discussion}: Follows the introduction; covers design ergonomics, materials, and cost-saving measures. \\
\addlinespace[2pt]
Boundary signals & Introduction of new documents or evaluation criteria \\
\addlinespace[2pt]
Segment duration & 120--300 seconds \\
\bottomrule
\end{tabular}
\end{table}

\section{Experimental Setup}
\label{sec:setup}

\textbf{Corpora.}\quad Table~\ref{tab:corpora} summarizes the six benchmarks,
covering 1,800 documents. We use their supplied transcripts, treating each sentence
or utterance as one indexed sentence. SIM tests topic changes introduced
by splicing unrelated meeting excerpts. YTSeg, AMI, and ICSI provide audio recordings and timestamped topic boundaries. For each evaluated configuration,
prompts and settings remain fixed across corpora. No model is fine-tuned for
segmentation or given labeled examples from the target corpora.

\textbf{Metrics.}\quad $P_k$ metric~\cite{beeferman1999statistical} measures how often sentences $i$ and $i+k$ share a predicted segment but not a ground-truth segment, or vice versa.
WindowDiff~\cite{pevzner2002critique} measures how often predicted and
ground-truth boundary counts differ within a window.
Both use $k=\max(1,\lfloor\bar{L}/2\rfloor)$, where $\bar{L}$ is mean
ground-truth segment length per document.
Boundary $F_1$~\cite{retkowski2024ytseg} is the harmonic mean of precision and recall.
We count a predicted boundary as correct if it is within two sentences of a
ground-truth boundary. Each boundary can be matched at most once.
Macro scores average documents within corpora, then corpora equally.

\textbf{Baselines.}\quad
\textsc{TextTiling}~\cite{hearst1997texttiling} detects changes in lexical
cohesion, while \textsc{BERT-TT}~\cite{solbiati2021unsupervised} uses
contextual embeddings to detect drops in similarity between adjacent
sentence blocks. The LLM baselines include a \textsc{TOC prompt}~\cite{freisinger2025toclm}, which generates section headings and
start indices; the recursive segmentation method of
Mackenzie et al.~\cite{mackenzie2025llmsegmentation}, which predicts boundary indices and
recursively divides long segments using fixed illustrative examples; and \textsc{LumberChunker}~\cite{duarte2024lumberchunker},
which iteratively identifies topic shifts, adapted here to sentence units.
\textsc{SegmentLLM}~\cite{segmentllm2024} uses a zero-shot prompt to group
consecutive sentences by topic, listing every sentence index in each group.
Segment boundaries are recovered from these groups after resolving gaps
and overlaps.
\textsc{Def-DTS}~\cite{defdts2025}
summarizes the preceding and following context of each utterance, classifies its intent, and predicts whether it starts a new topic. Generating these intermediate outputs for every utterance leads to substantial output-token usage on long transcripts. We apply Def-DTS to the full transcript, including speaker labels when available.

\begingroup\widowpenalty=10000
\textbf{Backbones.}\quad We evaluate six LLM backbones: Gemini-\allowbreak 3.1-\allowbreak Flash-\allowbreak Lite~\cite{gemini31flashlite2026},
GPT-\allowbreak 5.6-\allowbreak Terra~\cite{openai2026gpt56terra}, Qwen3.6-\allowbreak 27B~\cite{qwen36_2026},
Gemma-4-\allowbreak 26B-A4B~\cite{gemma4report2026}, Qwen3.5-\allowbreak 9B~\cite{qwen35_2026}, and
Qwen3.5-\allowbreak 4B~\cite{qwen35_2026}. We use temperature \(T{=}0.3\) for Flash-Lite and the open-weight models.
{Terra uses the API's default temperature because non-default values were not supported.}\par
\endgroup

\section{Results and Analysis}
\label{sec:results}
\label{sec:analysis}

\begin{table}[t]
  \centering
  \footnotesize
  \setlength{\tabcolsep}{4pt}
  \caption{Performance by backbone, averaged equally over the six corpora.
  Table~\ref{tab:main} averages LLM results over these backbones.
  \textbf{Bold} marks the better value.}
  \label{tab:backbones}
\vspace{3pt}
  \begin{tabular}{@{}lcccccc@{}}
    \toprule
    & \multicolumn{3}{c}{\textsc{SegmentLLM}} & \multicolumn{3}{c}{CGS (ours)} \\
    \cmidrule(lr){2-4} \cmidrule(l){5-7}
    Backbone & $P_k$ ($\downarrow$) & WD ($\downarrow$) & $F_1$ ($\uparrow$)
             & $P_k$ ($\downarrow$) & WD ($\downarrow$) & $F_1$ ($\uparrow$) \\
    \midrule
    Gemini-3.1-FL & .234 & .285 & .539 & \textbf{.177} & \textbf{.207} & \textbf{.557} \\
    GPT-5.6-Terra & .328 & .415 & .525 & \textbf{.181} & \textbf{.218} & \textbf{.550} \\
    Qwen3.6-27B & .241 & .278 & .495 & \textbf{.175} & \textbf{.208} & \textbf{.543} \\
    Gemma-4-26B & .411 & .605 & .363 & \textbf{.221} & \textbf{.296} & \textbf{.486} \\
    Qwen3.5-9B & .309 & .334 & .299 & \textbf{.230} & \textbf{.279} & \textbf{.464} \\
    Qwen3.5-4B & .319 & .356 & .248 & \textbf{.264} & \textbf{.337} & \textbf{.452} \\
    \bottomrule
  \end{tabular}
\end{table}

Table~\ref{tab:main} shows that, averaged over six backbones, CGS outperforms
all six published baselines on every corpus under all three metrics.
Figure~\ref{fig:motivation} summarizes section lengths from the same
predictions.
CGS also outperforms \textsc{SegmentLLM}, the strongest baseline in
Table~\ref{tab:main}, on all three macro metrics for each backbone
(Table~\ref{tab:backbones}). Across the six backbones, CGS completes segmentation after Stage~1 for $80$--$93\%$ of documents.

\begingroup\widowpenalty=10000
\textbf{Cost and token use.}\quad
Table~\ref{tab:cost} shows that CGS combines the lowest segmentation error with the second-lowest API cost. CGS generates about one-quarter as many output tokens as \textsc{SegmentLLM}, averaged across six backbones. On the two proprietary backbones, output tokens are priced six times higher than input tokens,
allowing the output savings to offset the additional input cost.\par
\endgroup

\begin{table}[!htbp]
\centering
\setlength{\tabcolsep}{4pt}
\caption{$P_k$ from Table~\ref{tab:main} and recorded token use (thousands per document). Per-document means are averaged across the six corpora and six backbones. Costs are calculated from recorded token usage at standard API rates and averaged over the two proprietary backbones (USD/1,000 documents). \textbf{Bold} and \underline{underline} mark the best and second-best values.}
\label{tab:cost}
\vspace{3pt}
\begin{tabular}{@{}lrrrr@{}}
\toprule
Method & $P_k\!\downarrow$ & Input (k)$\downarrow$ & Output (k)$\downarrow$ & Cost$\downarrow$ \\
\midrule
\textsc{LumberChunker} & .583 & 32.36 & 0.825 & 35.59 \\
Mackenzie et al. & .476 & 46.25 & \textbf{0.233} & 53.03 \\
\textsc{TOC prompt} & .497 & \underline{14.94} & \underline{0.306} & \textbf{17.17} \\
\textsc{Def-DTS} & .340 & 16.37 & 42.221 & 317.48 \\
\textsc{SegmentLLM} & \underline{.307} & \textbf{14.03} & 2.428 & 26.13 \\
\midrule
\textbf{CGS} & \textbf{.208} & 19.63 & 0.603 & \underline{21.87} \\
\bottomrule
\end{tabular}
\end{table}
\begin{samepage}
\textbf{CGS ablations.}\quad
Table~\ref{tab:ladder} reports ablations on Flash-Lite and Qwen3.6-27B, which achieve the strongest CGS performance among the evaluated proprietary and open-weight backbones, respectively.

\emph{Stage~1 only} (\textsc{Forced cues}) places boundaries at
the extracted topic openings and skips semantic fallback.
\emph{Stage~2 only} skips Stage~1 and segments every transcript directly.
These comparisons test whether either stage can replace the complete pipeline.
\emph{Cue indices only} keeps the two-stage procedure but asks Stage~1 to
report topic-opening sentence indices without quoting the corresponding
phrases. This tests whether generating the quotes helps boundary prediction.
CGS performs best among the variants in Table~\ref{tab:ladder} on both backbones.
\par\end{samepage}

\begin{samepage}
We also examine how much Stage-1 guidance improves Stage-2 segmentation. For the same documents routed to Stage~2, \emph{Count-only fallback} supplies the transcript and suggested segment-count range, but removes phase descriptions, boundary signals, and duration estimates. \emph{Transcript-only fallback} also removes the count range. Retaining the count range provides most of
the improvement over transcript-only fallback. Structural guidance reduces
$P_k$ further, with a larger effect on Qwen-27B.
\par\end{samepage}

\begin{table}[!htbp]
  \centering
  \small
  \setlength{\tabcolsep}{4pt}
  \caption{CGS ablations ($P_k\downarrow$, averaged equally over six corpora; $T{=}.3$). \textbf{Bold} marks the lowest displayed value.}
  \label{tab:ladder}
\vspace{3pt}
  \begin{tabular}{@{}lcc@{}}
    \toprule
    Configuration & Flash-Lite & Qwen-27B \\
    \midrule
    CGS & \textbf{.177} & \textbf{.175} \\
    \midrule
    Stage~1 only (\textsc{Forced cues}) & .197 & .187 \\
    Stage~2 only & .306 & .285 \\
    Cue indices only & .197 & .180 \\
    Count-only fallback & .192 & .183 \\
    Transcript-only fallback & .249 & .218 \\
    \bottomrule
  \end{tabular}
\end{table}

\begin{samepage}
\textbf{Boundary errors.}\quad Table~\ref{tab:boundary_errors} summarizes the correct, missed, and incorrect boundary counts used to calculate Boundary $F_1$ for each document. CGS detects about as many ground-truth boundaries as \textsc{SegmentLLM} and makes the fewest incorrect boundary predictions overall. Some baselines detect more ground-truth boundaries but also make many more incorrect predictions.
\par\end{samepage}

\begin{table}[!htbp]
\centering
\setlength{\tabcolsep}{7pt}
\caption{Counts are normalized to 100 ground-truth boundaries. Incorrect predictions can exceed 100 when a method predicts too many boundaries. Correct counts predictions that match a ground-truth boundary within two sentences. Each boundary is matched at most once. Missed reports how many ground-truth boundaries are not detected. Incorrect counts wrong boundary predictions. Values average six corpora and, for LLMs, six backbones.}
\label{tab:boundary_errors}
\vspace{3pt}
\begin{tabular}{@{}lrrr@{}}
\toprule
Method & Correct$\uparrow$ & Missed$\downarrow$ & Incorrect$\downarrow$ \\
\midrule
\textsc{TextTiling} & 61.6 & 38.4 & 1361.8 \\
\textsc{BERT-TT} & 13.3 & 86.7 & 71.6 \\
\textsc{LumberChunker} & \textbf{79.3} & \textbf{20.7} & 750.6 \\
Mackenzie et al. & 62.4 & 37.6 & 476.2 \\
\textsc{TOC prompt} & 41.4 & 58.6 & 262.4 \\
\textsc{Def-DTS} & 46.4 & 53.6 & 351.0 \\
\textsc{SegmentLLM} & 54.6 & 45.4 & 148.3 \\
\midrule
\textbf{CGS} & 55.5 & 44.5 & \textbf{58.2} \\
\bottomrule
\end{tabular}
\end{table}

\textbf{CGS retains its advantage on ASR transcripts.}\quad
Speech recognition errors and missing punctuation can make topic boundaries harder to locate. We use YTSeg's released Whisper-Large transcripts and generate AMI and ICSI transcripts with Whisper large-v3~\cite{radford2023whisper}. For AMI and ICSI, we compare the audio timestamp of each ground-truth boundary with Whisper's chunk start timestamps. A chunk starting at the boundary timestamp begins the new segment. If there is no exact timestamp match, the new segment begins with the first chunk after the boundary timestamp. Averaged across six backbones, CGS outperforms \textsc{SegmentLLM} on all metrics for all three corpora (Table~\ref{tab:asr}).
\par

\begin{table}[!ht]
\centering
\caption{ASR segmentation performance, averaged over six backbones. Sentence divisions differ between the original and ASR transcripts.}
\label{tab:asr}
\vspace{3pt}
{\setlength{\tabcolsep}{7pt}
\providecommand{\asrscorecell}[1]{{$#1$}}
\begin{tabular}{@{}llrrr@{}}
\toprule
Corpus & Method & $P_k\downarrow$ & WD$\downarrow$ & $F_1\uparrow$ \\
\midrule
AMI & SegmentLLM & \asrscorecell{.278} & \asrscorecell{.359} & \asrscorecell{.420} \\
 & \textbf{CGS} & \asrscorecell{\mathbf{.202}} & \asrscorecell{\mathbf{.263}} & \asrscorecell{\mathbf{.512}} \\
\addlinespace[3pt]
ICSI & SegmentLLM & \asrscorecell{.310} & \asrscorecell{.391} & \asrscorecell{.355} \\
 & \textbf{CGS} & \asrscorecell{\mathbf{.244}} & \asrscorecell{\mathbf{.307}} & \asrscorecell{\mathbf{.456}} \\
\addlinespace[3pt]
YTSeg & SegmentLLM & \asrscorecell{.277} & \asrscorecell{.335} & \asrscorecell{.577} \\
 & \textbf{CGS} & \asrscorecell{\mathbf{.253}} & \asrscorecell{\mathbf{.286}} & \asrscorecell{\mathbf{.592}} \\
\bottomrule
\end{tabular}

}
\end{table}

\begin{samepage}
\section{Conclusion}
\label{sec:discussion}

CGS grounds topic boundaries in quoted transition cues and invokes structure-guided semantic segmentation when these cues are insufficient. Averaged across the six corpora, it outperforms the evaluated baselines on all metrics with each of the six backbones. Ablations show the benefit of combining cue extraction with selective semantic fallback and identify segment-count guidance as an important contributor to fallback performance. CGS also retains its performance advantage on noisy ASR transcripts, supporting robustness to speech recognition errors. By combining accurate segmentation with low API cost, CGS offers a practical approach to segmenting spoken documents without task-specific training.
\par\end{samepage}

\clearpage
\section{Acknowledgments}

This work was supported by (1) the National Research Foundation of Korea (NRF) grant funded by the Korea government (MSIT) (RS-2026-25477522), (2) Institute of Information \& communications Technology Planning \& Evaluation (IITP) grant funded by the Korea government (MSIT) [NO.RS-2021-II211343, Artificial Intelligence Graduate School Program (Seoul National University)], and (3) the Starting growth Technological R\&D Program (RS-2024-00506994) funded by the Ministry of SMEs and Startups (MSS, Korea).

\section{Compliance with Ethical Standards}

This study uses publicly released research corpora under their respective terms and involves no new human-subject data collection.
\urlstyle{same}
\bibliographystyle{IEEEbib}
\bibliography{refs}

\end{document}